\documentclass[letterpaper, 10 pt, conference]{ieeeconf}  % Comment this line out if you need a4paper

\IEEEoverridecommandlockouts                              % This command is only needed if 
\usepackage{graphics} % for pdf, bitmapped graphics files
\usepackage{epsfig} % for postscript graphics files
\usepackage{mathptmx} % assumes new font selection scheme installed
\usepackage{times} % assumes new font selection scheme installed
\usepackage{amsmath} % assumes amsmath package installed
\usepackage{amssymb}  % assumes amsmath package installed

\usepackage{threeparttable}
\usepackage{textcomp}
\usepackage{siunitx}
\usepackage{makecell}
\usepackage{adjustbox}
\usepackage{diagbox}
\usepackage{booktabs, multirow}
\usepackage[utf8]{inputenc}

\usepackage{url}

\usepackage{xcolor}
\usepackage{graphicx}
\usepackage{tikz}
\usepackage{listings}                       % Source code
\usepackage{algorithm, algorithmic}                      % Pseudo Code
\usetikzlibrary{backgrounds}

\usepackage[noadjust]{cite}

\usepackage{gensymb}

\title{\LARGE \bf
Seg2Grasp: A Robust Modular Suction Grasping in Bin Picking}

\author{Hye-Jung Yoon$^{1*}$ Juno Kim$^{1*}$ Yesol Park$^{1*}$ Jun-Ki Lee$^{2}$ Byoung-Tak Zhang$^{1,2,3}$%
    \thanks{*Authors have equal contributions}% <-this % stops a space
    \thanks{$^{1}$Interdisciplinary Program in AI, Seoul National University}%
    \thanks{$^{2}$Artificial Intelligence Institute, Seoul National University}%
    \thanks{$^{3}$Department of Computer Science, Seoul National University}%
    \thanks{This work was partly supported by Institute of Information \& Communications Technology Planning \& Evaluation (IITP) grants funded by the Korean government (MSIT) [RS-2021-II211343, Artificial Intelligence Graduate School Program (Seoul National University)] and (RS-2021-II212068-AIHub/10\%, RS-2021-II211343-GSAI/15\%, 2022-0-00951-LBA/15\%, 2022-0-00953-PICA/20\%), NRF (RS-2024-00353991/20\%, RS-2023-00274280/10\%), and KEIT (RS-2024-00423940/10\%).
    }%
}

\begin{document}

\maketitle
\thispagestyle{empty}
\pagestyle{empty}

%%%%%%%%%%%%%%%%%%%%%%%%%%%%%%%%%%%%%%%%%%%%%%%%%%%%%%%%%%%%%%%%%%%%%%%%%%%%%%%%
\begin{abstract}

% Current bin picking methods predominantly rely on end-to-end learning-based strategies to identify optimal picking points. However, these approaches struggle in environments with unfamiliar or unseen objects. To address these limitations, we introduce Seg2Grasp, a modular pipeline designed for robust suction grasping in dynamic, cluttered bin scenarios without requiring controlled environments or prior object knowledge. Seg2Grasp follows a \textit{Segmentation-Grasping-Classification} approach: the \textit{Segmentation} module generates class-agnostic mask proposals from RGB-D images, effectively delineating objects regardless of their size, shape, or orientation; the \textit{Grasping} module uses these proposals and surface normals to identify optimal suction points; and the \textit{Classification} module employs open-vocabulary classification to adapt to a wide range of objects with precision. Seg2Grasp outperforms existing methods in real-world robotic experiments, achieving higher grasp success rates and better adaptability to diverse and complex object configurations, demonstrating its effectiveness as a versatile solution for automated picking operations.

Current bin picking methods that rely heavily on end-to-end learning often falter when confronted with unfamiliar or complex objects in unstructured environments. To overcome these limitations, we introduce Seg2Grasp, a modular pipeline designed for robust suction grasping in dynamic and cluttered bin scenarios. Seg2Grasp is built on a three-step process: \textit{Segmentation}, \textit{Grasping}, and \textit{Classification}. The \textit{Segmentation} module employs a Transformer-based model to generate class-agnostic object masks from RGB-D images, ensuring accurate detection across various conditions. The \textit{Grasping} module uses surface normals and mask proposals to determine the optimal suction points, enhancing grasp success. Finally, the \textit{Classification} module leverages fine-tuned open-vocabulary Mask-CLIP for precise object identification, enabling versatile handling of diverse objects. Real-world robotic experiments demonstrate that Seg2Grasp outperforms existing methods in success rates and adaptability, establishing it as a powerful tool for automated bin picking in industrial settings.

\end{abstract}

%%%%%%%%%%%%%%%%%%%%%%%%%%%%%%%%%%%%%%%%%%%%%%%%%%%%%%%%%%%%%%%%%%%%%%%%%%%%%%%%

\section{INTRODUCTION}

In the field of industrial automation, bin picking is a critical yet challenging task, especially when applied to environments that are dynamic and contain unknown objects. Such settings are characterized by a variety of unpredictable factors, including fluctuating lighting, different camera viewpoints, and the presence of objects that the system has not previously encountered. These challenges demand a solution that is both precise and adaptable, capable of handling the complexities of real-world industrial scenarios.
 
Existing bin picking systems predominantly utilize end-to-end learning-based methods~\cite{mahler2017dex, mahler2018dex, dexnet40, cao2021suctionnet, zeng2022robotic}, which aim to directly map sensory inputs to outputs. Although these approaches have shown success in controlled environments, they frequently fall short in terms of adaptability—the ability to function effectively across diverse conditions—and robustness, which refers to maintaining performance despite variations in the environment. The relatively simplistic nature of these end-to-end methods limits their generalization capabilities, particularly when faced with novel objects or unfamiliar configurations.

\begin{figure} [t!]
\centering
\includegraphics[width=0.44\textwidth]{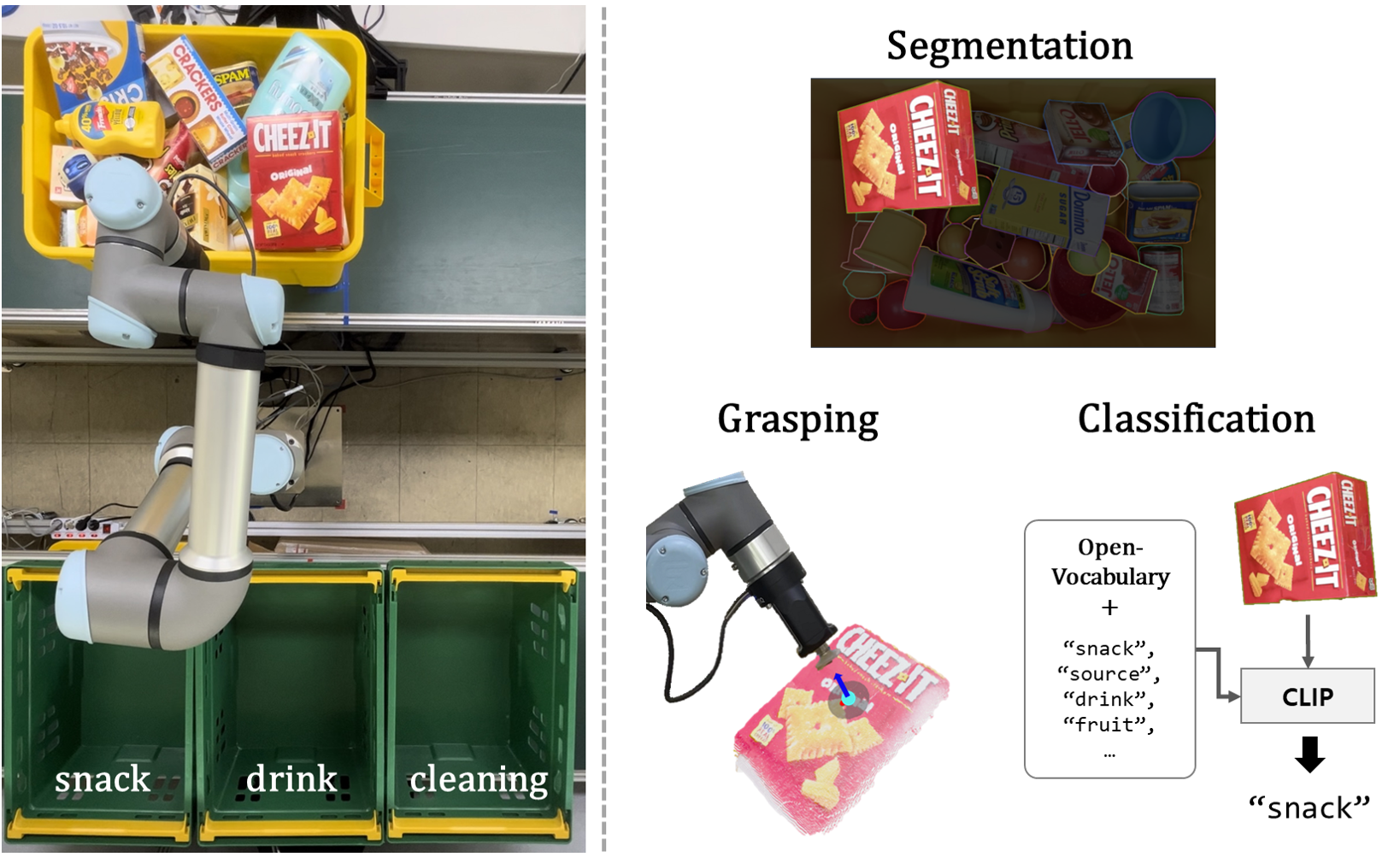}
\vspace{-1mm}
\caption{ \textbf{Illustration of proposed system.} Our bin picking system can segment variously shaped, class-agnostic objects in a dynamic environment and proceed with grasping and classifying them.}

\label{introduction}
\vspace{-5mm}
\end{figure}

To address these limitations, we propose Seg2Grasp, a novel pipeline designed to enhance the robustness and adaptability of suction-based bin picking systems in dynamic environments. Unlike traditional end-to-end models, Seg2Grasp leverages a modular architecture comprising three core components: \textit{Segmentation, Grasping,} and \textit{Classification}. This modular design, depicted in Fig.~\ref{introduction}, allows each component to specialize in a specific task, providing greater flexibility and improved performance across a wider range of scenarios.

The Segmentation module utilizes a Transformer-based model to generate class-agnostic object masks from depth-weighted RGB images. This method enables accurate object segmentation across diverse conditions, overcoming one of the primary challenges in bin picking—reliable detection and segmentation of objects regardless of environmental variability.

These object masks are then passed to the Grasping module, which, combined with surface normal data, identifies optimal grasping points. This process significantly improves the precision of object manipulation, enabling the system to successfully grasp a wide range of objects with varying geometries and orientations.

Finally, the Classification module employs an open-vocabulary classification system to categorize a wide array of objects. This capability allows the system to adapt to and handle objects without prior explicit knowledge of each object, greatly expanding the operational versatility of the robotic system.

\begin{figure*} [t!]
\centering
{\includegraphics[width=\textwidth] {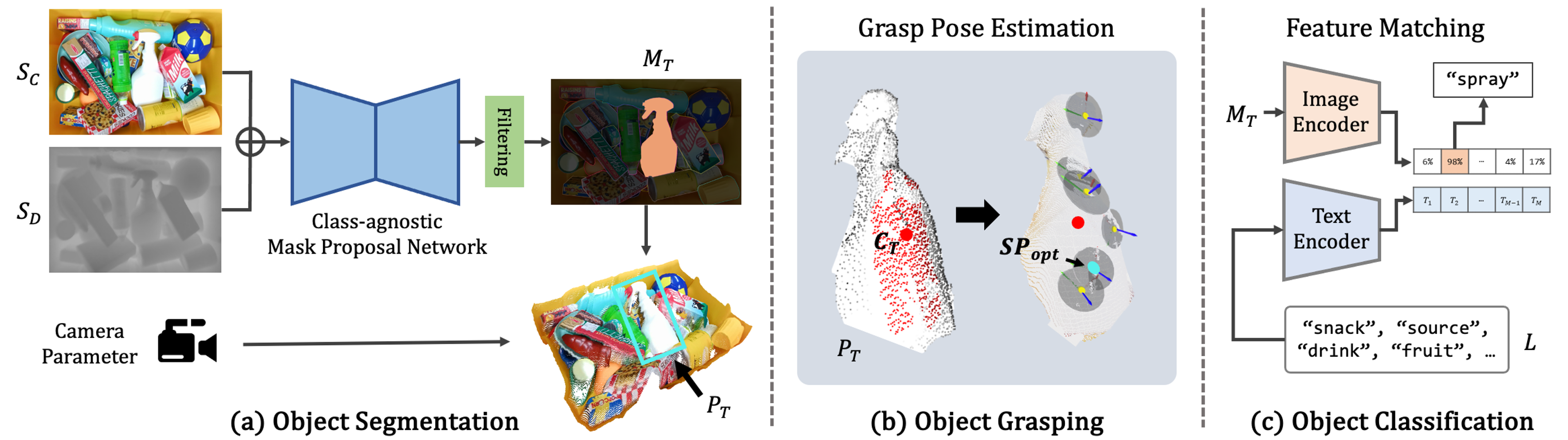}} %[width=165mm, height=50mm] 
\vspace{-5mm}
\caption{\textbf{Overview of our modular bin picking system.} The system comprises three modules: class-agnostic object segmentation, object grasp pose estimation, and open-vocabulary object classification. The black arrow indicates the flow of operations within the entire system.}
\label{task_overview}
\vspace{-5mm}
\end{figure*}

To validate the effectiveness of Seg2Grasp, we conducted a series of robotic experiments in dynamic environments, utilizing various bins and camera configurations to simulate realistic bin picking scenarios. Our approach consistently outperformed existing end-to-end methods~\cite{cao2021suctionnet, dexnet40} in terms of success rates, particularly in unstructured and dynamic environments.

In summary, Seg2Grasp addresses the significant gaps in adaptability and robustness found in current bin picking technologies. By integrating specialized modules for segmentation, grasping, and classification, Seg2Grasp not only enhances performance in complex environments but also sets a new benchmark for the potential of modular approaches in industrial automation.

\section{Related Work}

In robotics, bin picking in unstructured environments is a complex task that requires robust object recognition, accurate grasp pose estimation, and flexible object classification. This section reviews the recent advances in these areas, which are essential to the development of our Seg2Grasp framework.

\subsection{Class-agnostic Object Instance Segmentation}
Class-agnostic Object Instance Segmentation is critical for enabling robots to identify and segment objects that have not been previously encountered. Traditional segmentation methods often struggle in unstructured and cluttered environments~\cite{uckermann2012real, mohammed2022review}. Unlike category-based instance segmentation, which focuses on known categories~\cite{zhu2020deformable, cheng2022masked, tian2020conditional}, Class-agnostic Object Instance Segmentation aims to generalize segmentation to arbitrary objects~\cite{xie2020best, xie2021unseen, xiang2021learning}.

Recent advancements include UOAIS-Net~\cite{back2022unseen}, which addresses occlusion and amodal perception in robotic manipulation but faces challenges with occluded areas, leading to potential misinterpretations in cluttered scenarios. Another approach, MSMFormer~\cite{lu2022mean}, improves segmentation accuracy by integrating a clustering method with Mask2Former~\cite{cheng2022masked}. However, this comes at the cost of increased computational requirements, highlighting a precision-efficiency trade-off.

Our approach adopts Mask2Former for its robust segmentation capabilities, particularly its ability to maintain high accuracy without the need for clustering. This selection strikes a balance between efficiency and performance, making it suitable for the dynamic environments typical of bin picking tasks.

\subsection{Grasp Pose Estimation}
Grasp pose estimation is fundamental in robotic manipulation, where accurate identification of grasp points is necessary for successful object handling. This process typically relies on RGB-D and point cloud data~\cite{9784795}. There are two main approaches: learning-free methods~\cite{spenrath2017gripping} and learning-based methods~\cite{dong2019ppr, li2022sim, mahler2018dex, cao2021suctionnet, zhang2021cnn, zeng2022robotic}.

Learning-free approaches often depend on CAD models to determine grasp candidates for recognized objects~\cite{spenrath2017gripping, schillinger2023model}. On the other hand, learning-based methods, particularly those utilizing deep learning, have gained prominence due to their improved accuracy and reliability in controlled environments~\cite{chen2020survey}. These models are generally trained on depth or RGB-D datasets~\cite{dexnet40, cao2021suctionnet}, which allows for precise and consistent grasping. However, their performance tends to degrade in unstructured environments where the conditions differ from the training data.

To address the limitations of traditional methods, our grasping algorithm is specifically designed to maintain robustness in diverse and unpredictable environments. This approach ensures that grasp performance remains effective even under the variable conditions typical of real-world industrial settings, thus filling a significant gap in current robotic manipulation strategies.

\subsection{Open-Vocabulary Classification}
Open-vocabulary classification represents a significant advancement in image classification, allowing systems to recognize a broader range of visual concepts beyond fixed label sets, enhanced by natural language processing. Language-Image Pre-training models, such as CLIP~\cite{radford2021learning}, have been instrumental in bridging the gap between visual data and textual descriptions.

Despite its strengths, CLIP and similar models often experience reduced performance when applied to images that deviate from their training distributions. OVSeg~\cite{liang2023open} addresses this issue by extracting CLIP features directly from segmentation masks, excluding background elements to enhance accuracy. This refined approach improves the model’s performance, particularly when dealing with images significantly different from those in the training set.

In our Seg2Grasp framework, we utilize an open-vocabulary classification system enhanced by Mask-CLIP, which enables effective object recognition across a wide variety of scenarios. This flexibility further expands the operational capabilities of the system in dynamic bin picking environments.

\begin{figure} [t!]
\centering
{\includegraphics[width=0.48\textwidth]{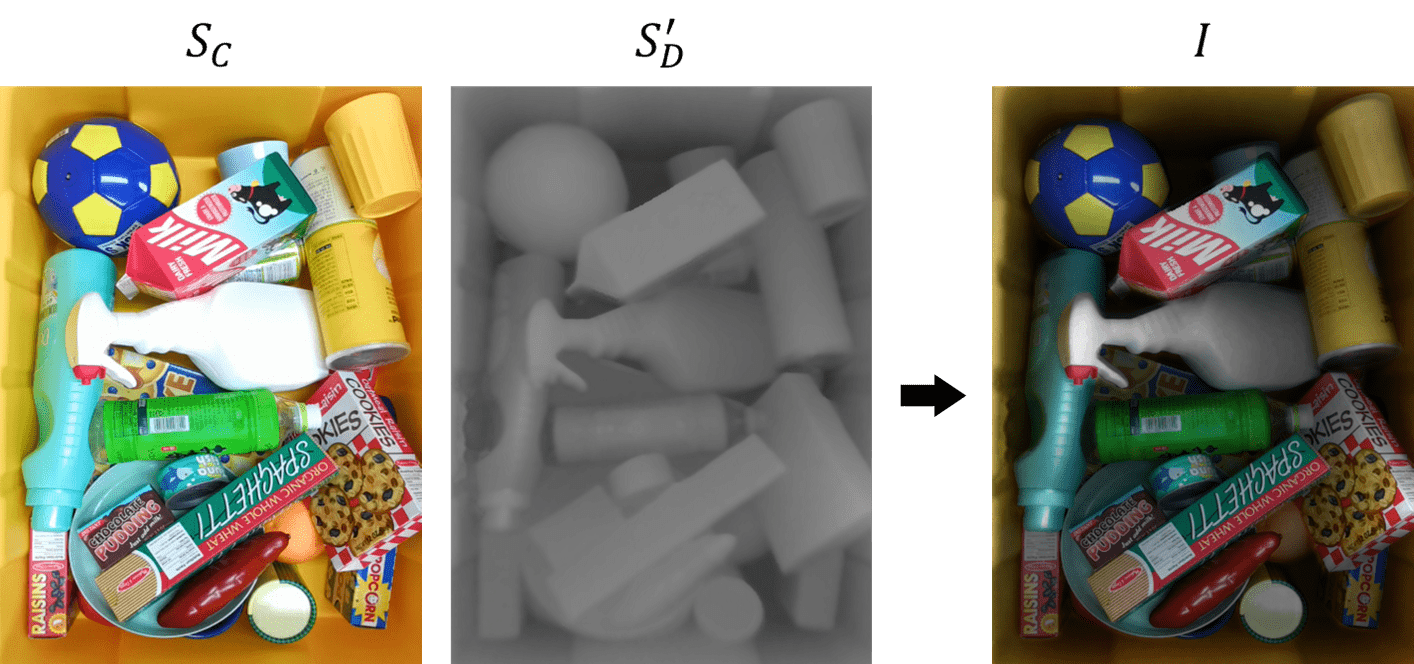}}
\vspace{-5mm}
\caption{\textbf{Mask proposal network input.} The input \(I\) to the mask proposal network consists of a fusion of the color image $S_{C}$ and the inverted depth image $S'_{D}$.}
\label{input_image}
\vspace{-5mm}
\end{figure}

\section{Problem Statement}
In the context of bin picking, our primary objective is to develop a system capable of accurately predicting a set of feasible grasp poses and their corresponding object labels. This system must effectively transfer arbitrary objects from a source bin to a target bin, even in unstructured and cluttered environments. Achieving this goal requires a comprehensive understanding of the scene, including precise segmentation, grasp pose estimation, and object classification.

Given a scene image \( S \), the task is to segment the image into distinct objects, represented by a set of masks \( \{M_i \mid 1 \leq i \leq N\} \), where \( N \) is the number of objects identified in the scene. Each mask \( M_i \) corresponds to an individual object, facilitating further analysis. The target object in the scene is denoted by \( T \).

The grasp configuration for the target object \( T \) is defined as \( g = (SP_{opt}, NV_{opt}) \), where \( SP_{opt} \) represents the optimal suction point on the object, and \( NV_{opt} \) is the associated normal vector at that point. Additionally, each object is assigned a label \( l \), which indicates its categorical identity and is essential for its correct placement.

The core challenge we address is to develop a predictive model \( f \) that, for each input mask \( M_i \), accurately maps to both a grasp configuration \( g_i = (SP_{opt,i}, NV_{opt,i}) \) and an object label \( l_i \). This mapping is expressed as:
\[
f: M_i \longmapsto (g_i, l_i), \quad \text{for } i = 1, 2, \dots, N.
\]
This formulation is critical for determining not only the optimal grasping approach but also for understanding the object's identity, ensuring its correct transfer and placement.

To solve this problem, we propose a three-step method:
\begin{enumerate}
    \item \textbf{Object Segmentation:} A Transformer-based model processes the scene image \( S \) to generate class-agnostic object masks, producing the set \( \{M_i \mid 1 \leq i \leq N\} \). The target object \( T \) is selected from this set.
    \item \textbf{Object Grasping:} For the selected target object \( T \), surface normals are used to calculate the optimal grasp pose \( g = (SP_{opt}, NV_{opt}) \), ensuring the most suitable suction point is identified.
    \item \textbf{Object Classification:} An open-vocabulary classification module assigns a label \( l \) to each object, facilitating accurate identification and enabling precise placement.
\end{enumerate}

Our modular approach, which separates segmentation, grasping, and classification into distinct yet interconnected components, offers enhanced adaptability and robustness. This flexibility is particularly advantageous in dynamic and variable environments, where conditions may differ significantly from the controlled settings typically assumed by other methods.

\section{METHOD}
Seg2Grasp is structured into three main modules: (1) \textit{Object Segmentation}, (2) \textit{Object Grasping}, and (3) \textit{Object Classification}. The overall framework is illustrated in Fig.~\ref{task_overview}
  
\subsection{Object Segmentation} The object segmentation module processes the scene image $S$, and distinguishes individual objects within the scene. The target object selected for grasping is denoted as $T$.

\textbf{Input Preparation.} The input to the network, denoted as \(I\), is constructed by combining the RGB image \(S_C\) with a modified depth image \(S'_D\). To enhance the depth information, the depth image \(S_D\) is normalized and inverted:

\begin{equation}
S'_D = 1 - \left( \frac{S_D - S_{D_{\text{min}}}}{S_{D_{\text{max}}} - S_{D_{\text{min}}}} \right),
\end{equation}

where \(S_{D_{\text{min}}}\) and \(S_{D_{\text{max}}}\) are the minimum and maximum values in \(S_D\). This inversion highlights closer objects, improving the contrast between objects at different depths. The final input \(I\) is the product of the color image \(S_C\) and the inverted depth image \(S'_D\), which enhances object differentiation. This process is shown in Fig.~\ref{input_image}

\begin{algorithm} [t!]
\caption{Mask Filtering}
\label{filtering}
\begin{algorithmic}[1]
\REQUIRE Set of masks \(M\)
\ENSURE Target object \(T\), Largest planar area \(A_T\)

\FOR{each mask \(M_i\) in \(M\)}
    \STATE Initialize: \(LP_{i} \gets\) null, Best inlier count \(In_{\text{best},i} \gets 0\)
    \FOR{iteration \(k\)}
        \STATE Select a random set \(X\) from \(M_i\)
        \STATE Estimate plane \(Pl_k\) from \(X\)
        \STATE Determine inliers \(In_k\): Points in \(M_i\) fitting \(Pl_k\) within tolerance \(\epsilon\)
        \IF{\( |In_k| > In_{\text{best},i} \)}
            \STATE \(LP_{i} \gets Pl_k\), \(In_{\text{best},i} \gets |In_k|\)
        \ENDIF
    \ENDFOR
    \STATE Calculate centroid \(C_i = (x_i, y_i, z_i)\) of \(In_{\text{best},i}\)
\ENDFOR

\STATE \(T = \arg\max_{i}(z_i)\) % Identify \(T\) with the highest \(z\)-value among \(C_i\)
\STATE \(A_T = \text{Area of } LP \text{ for } T\)

\RETURN \(T\), \(A_T\)
\end{algorithmic}
\end{algorithm}

\textbf{Mask Proposal.} We developed a model for proposing class-agnostic instance masks using the standard Mask Transformer model~\cite{cheng2022masked}. The model outputs a set of masks \(M = \{M_i \mid 1 \leq i \leq 100\}\) for each input image, where each mask captures an individual object. By using the fused input \(I\) and focusing on class-agnostic outputs, the model generalizes well across diverse objects.

\textbf{Mask Filtering.}
 To identify the target object \(T\), we refine the set of generated masks \(M\) using a filtering process. This process leverages the RANSAC algorithm to estimate the largest planar area (\(LP_i\)) within each mask \(M_i\). For each mask, RANSAC iteratively selects a subset of points to estimate a plane and identifies inliers—points that fit the plane within a specified tolerance. The plane with the highest inlier count is selected, and the centroid \(C_i = (x_i, y_i, z_i)\) of these inliers is computed. The mask with the highest \(z\)-coordinate centroid is then chosen as the target object \(T\), ensuring that the most elevated object in the scene, and thus the most accessible for grasping, is selected. The complete mask filtering process is detailed in Alg.~\ref{filtering}.

\begin{figure} [t!]
% \vspace{-2mm}
\centering
{\includegraphics[width=0.47\textwidth]{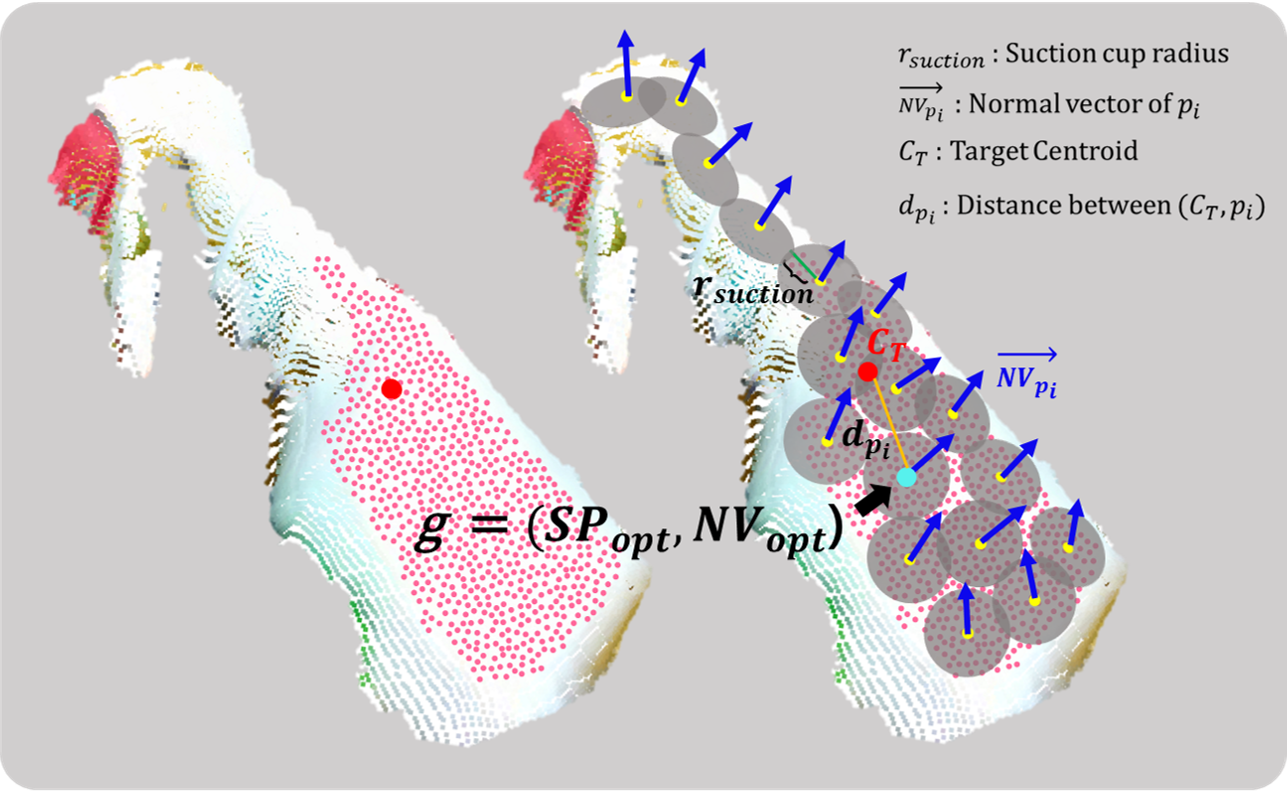}}
\vspace{-2mm}
\caption{\textbf{Grasp pose estimation.} The cyan dot represents the optimal grasp point on the target object, identified by the highest final score (detailed in Alg.~\ref{alg:pose_estimation}).
}
\label{Grasp}
\vspace{-3mm}
\end{figure}

\subsection{Object Grasping}
The object grasping module processes the selected target object \(T\) to determine the optimal grasp point \(g = (SP_{opt}, NV_{opt})\), as depicted in Fig.~\ref{Grasp}. This process, detailed in Alg.~\ref{alg:pose_estimation}, utilizes surface normals derived from the object to identify the most suitable suction pose for grasping.

Our approach differs from traditional methods~\cite{cao2021suctionnet, dexnet40}, which often depend on extensive training data and specific environmental setups. Instead, we rely solely on RGB-D imagery to compute the optimal suction point, providing greater flexibility across various objects and scenarios. This is particularly advantageous when using suction grippers, which can attach to suitable surface points on diverse objects.

\textbf{Suction Point Evaluation.} The optimal suction point is determined by analyzing the preprocessed point cloud \(P_{T}\). The process involves navigating through \(P_{T}\) to identify potential suction areas, with the region of interest adjusted to match the suction cup's dimensions. Each potential grasp point is evaluated based on its alignment with the normal vector \(NV_{T}\) of the central point, considering a tolerance angle \(\phi\).

To determine the most suitable suction point \(g = (SP_{opt}, NV_{opt})\), each candidate site is assessed using three metrics: surface angle \(\theta\), distance \(d\) from the centroid \(C_{T}\), and the count of graspable points \(\mathcal{G}\). These metrics are normalized and combined into a composite score that prioritizes angular alignment, proximity, and point count:

\begin{equation}
\text{FinalScore}(g) = w_\theta S_\theta(\theta) + w_d S_d(d) + w_g S_g(|\mathcal{G}|)
\end{equation}

Here, \(S_\theta\), \(S_d\), and \(S_g\) are the normalized score functions for each metric, with \(w_\theta\), \(w_d\), and \(w_g\) as their respective weights, satisfying the condition \(w_\theta + w_d + w_g = 1\).

\begin{algorithm}[t!]
\caption{Pose Estimation for Optimal Suction Point}
\label{alg:pose_estimation}
\begin{algorithmic}[1]
\REQUIRE Point cloud \(P_{T}\), 2D mask \(M_{T}\), centroid \(C_{T}\), suction radius \(r_{suction}\), angle tolerance \(\phi\), min grasped points \(\psi\), angle threshold \(\delta\)
\ENSURE Optimal suction point \(SP_{opt}\), normal vector \(NV_{opt}\) at \(SP_{opt}\)

\STATE \textit{Preprocessing:}  
\STATE Filter and downsample \(P_{T}\) using \(M_{T}\).
\STATE Calculate surface normals \(NV_{T}\) and angles \(\Theta_{T}\).

\STATE \textit{Point Evaluation:}
\FOR{each point \(p\) in \(P_{T}\)}
    \IF{angle \(\Theta_{T}(p)\) $>$ \(\delta\)}
        \STATE Define region \(R_{near}\) around \(p\) within \(r_{suction}\).
            \STATE Filter points in \(R_{near}\) with normals within \(\phi\) of \(NV_T(p)\).
        \IF{count of filtered points \(> \psi\)}
            \STATE Determine plane normal \(N_{plane}\) for these points.
            \STATE Aggregate \(N_{plane}\), its angle, distance to \(C_{T}\), and inlier count.
        \ENDIF
    \ENDIF
\ENDFOR

\STATE Aggregate and weight selection criteria \(S_\theta, S_d, S_g\) with weights \(w_\theta, w_d, w_g\).
\STATE Identify \(SP_{opt}\) and \(NV_{opt}\) using weighted criteria.
\RETURN \(SP_{opt}, NV_{opt}\)
\end{algorithmic}
\end{algorithm}

\subsection{Object Classification}
The classification module operates in parallel with the grasping mechanism, identifying the target object \(T\) and assigning it a corresponding label \(l\), thereby enabling precise placement.

\textbf{Preparation for CLIP.}
We employ a fine-tuned Mask-CLIP model for open-vocabulary object classification, following the methodology outlined in previous research~\cite{liang2023open}. The model processes the mask proposals generated earlier, with a prediction branch specifically tailored for masked inputs. A CLIP-based branch then computes the class probabilities \(\hat{p}_{i,k}\) for each mask. A key challenge arises from the fact that CLIP, being trained primarily on images with natural backgrounds, exhibits reduced effectiveness when working with masked inputs that contain large areas of blank space.

\textbf{Fine-Tuning Mask-CLIP.}
Masked images, when tokenized for CLIP, often lead to zero tokens due to the extensive blank areas, indicating a domain shift. To address this, we apply mask prompt tuning. This involves enhancing the tensorized masked images with learnable prompt tokens derived from a binary mask, helping to preserve essential boundary information and improve classification accuracy.

\textbf{Feature Matching.}
After fine-tuning the Mask-CLIP model, we measure the cosine similarity between the features of the masked images and the text descriptors of the categories. This approach enables precise object classification by effectively matching the visual features to the corresponding textual descriptions.

\section{EXPERIMENTS}
This section presents the experimental setup and results, which demonstrate the efficacy of the proposed method in addressing our problem context.

\subsection{Implementation Detail}

\subsubsection{Segmentation Module}
Accurate object segmentation within cluttered bin environments, especially when encountering class-agnostic items, requires a model that generalizes well across different scenarios. This capability is crucial for successful bin picking tasks.

\textbf{Dataset.} Given the scarcity of comprehensive real-world datasets for bin picking, we utilized the UOAIS-SIM dataset~\cite{back2022unseen}, which includes 50,000 photorealistic RGB-D images of objects in bin settings. This dataset bridges the gap between simulation and reality, offering extensive exposure to various object scenarios critical for training robust models.

%%%%%%%%%%%%%수식 있는 버전

% \textbf{Training.} To optimize our segmentation module, we incorporate the Tversky loss function, designed to address the challenges of class imbalance and to focus on class-agnostic object detection. The Tversky loss is given by:
% $L_{\text{Tversky}} = \frac{\sum_{\text{pixels}} p_{g} \cdot p_{p}}{\sum_{\text{pixels}} p_{g} \cdot p_{p} + \alpha \sum_{\text{pixels}} (1 - p_{g}) \cdot p_{p} + \beta \sum_{\text{pixels}} p_{g} \cdot (1 - p_{p})}$,
% where $p_{g}$ and $p_{p}$ represent the ground truth and predicted probabilities for each pixel, respectively, and $\alpha = 0.7$, $\beta = 0.3$ are chosen to fine-tune the balance between false positives and false negatives, with a greater emphasis on minimizing false negatives.

% The final loss combines Tversky loss with a classification loss to ensure both accurate segmentation and classification:
% \[
% L_{\text{final}} = L_{\text{Tversky}} + \lambda_{\text{cls}} L_{\text{cls}},
% \]
% where \(\lambda_{\text{cls}} = 1.5\). This combined loss is optimized using the AdamW optimizer with a learning rate of \(1e-4\), and training is conducted over five epochs.

%%%%%%%%%%%%%수식 없는 버전

\textbf{Training.} We optimized our segmentation module using the Tversky loss function, which addresses class imbalance by focusing on class-agnostic object detection. The final loss combines Tversky loss and a classification loss component to ensure comprehensive learning. Training was conducted on the UOAIS-SIM dataset using the AdamW optimizer with a learning rate of $1e-4$ and a batch size of 4 over five epochs.

\begin{figure} [t!]
% \vspace{-2mm}
\centering
{\includegraphics[width=0.48\textwidth]{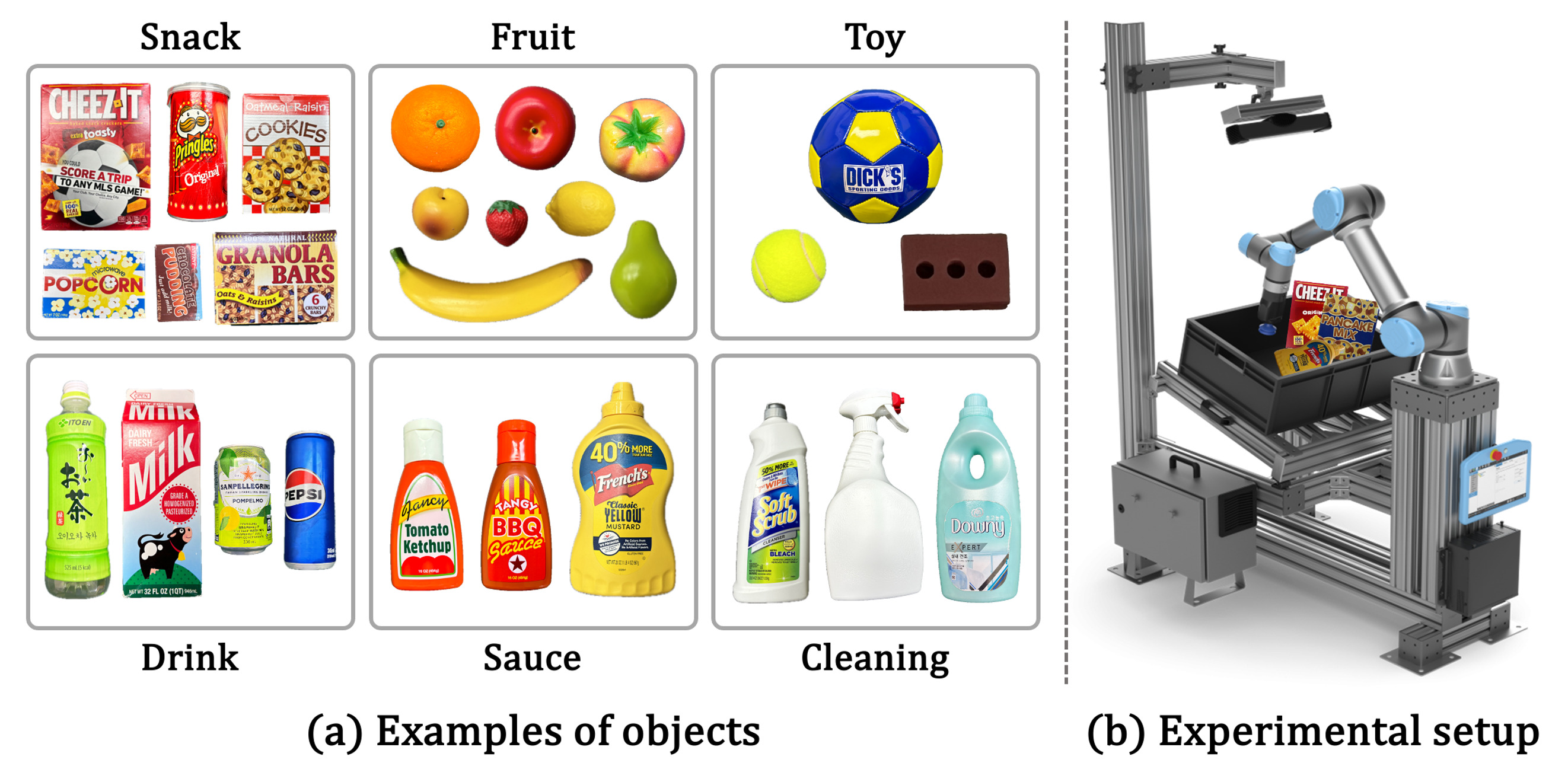}}
\vspace{-5mm}
\caption{\textbf{Experimental setup.} (a) Examples of objects used in the bin picking experiments. (b) Experimental environment using the UR5e robot.
}
\label{setting_fig}
\vspace{-5mm}
\end{figure}

\subsubsection{Classification Module}
To enhance the accuracy of open-vocabulary object classification with mask-only inputs, fine-tuning is imperative. This adjustment ensures the system can effectively recognize objects across a wide range of categories.

\textbf{Dataset.}
For our experiment, we constructed a product database using the `Product Image Dataset'~\cite{AI-Hub}. Given the variability in object orientations within bins, we developed a `mask-category' dataset. This dataset encompasses 720,000 product mask images across 53 main categories (e.g., snack, drink, dairy), generated from photographs of 10,000 objects captured from 72 different angles. This comprehensive approach allows our model to better recognize objects from any orientation, enhancing the robustness of our classification module.

\textbf{Training.} The OpenCLIP framework~\cite{ilharco_gabriel_2021_5143773} was fine-tuned using our specialized dataset. We used 90\% of the data for training and 10\% for evaluation, ensuring balanced exposure across all categories. The model was optimized for accurate classification of masked images, with training conducted over 10 epochs using a ViT-L/14 CLIP variant.

\subsection{Real-robot Experiments}
To demonstrate the robustness of our system, we designed three distinct experimental setups: 1) \textit{Optimal Conditions}, to establish a baseline for performance; 2) \textit{Varying Camera Parameters}, to examine how changes in visual input affect system performance; and 3) \textit{Different Bin Environments}, to assess the system's adaptability to changes in the surrounding environment.

\textbf{Experimental Setup.}
For our experiments, we selected a diverse range of objects, including boxes, cylinders, spheres, and various irregular shapes, as shown in Fig.~\ref{setting_fig}(a). This selection ensured a comprehensive evaluation of our system's adaptability. A UR5e robotic arm equipped with a Robotiq AirPick Vacuum Gripper was used for object manipulation, while RGB-D data was captured using an Azure Kinect DK Camera, strategically positioned at an elevated angle, as shown in Fig.~\ref{setting_fig}(b). The experiment was designed to focus on objects compatible with the vacuum gripper, allowing us to optimize the evaluation of the system's performance in real-world scenarios.

\begin{figure*} [t!]
\centering
{\includegraphics[width=0.98\textwidth]{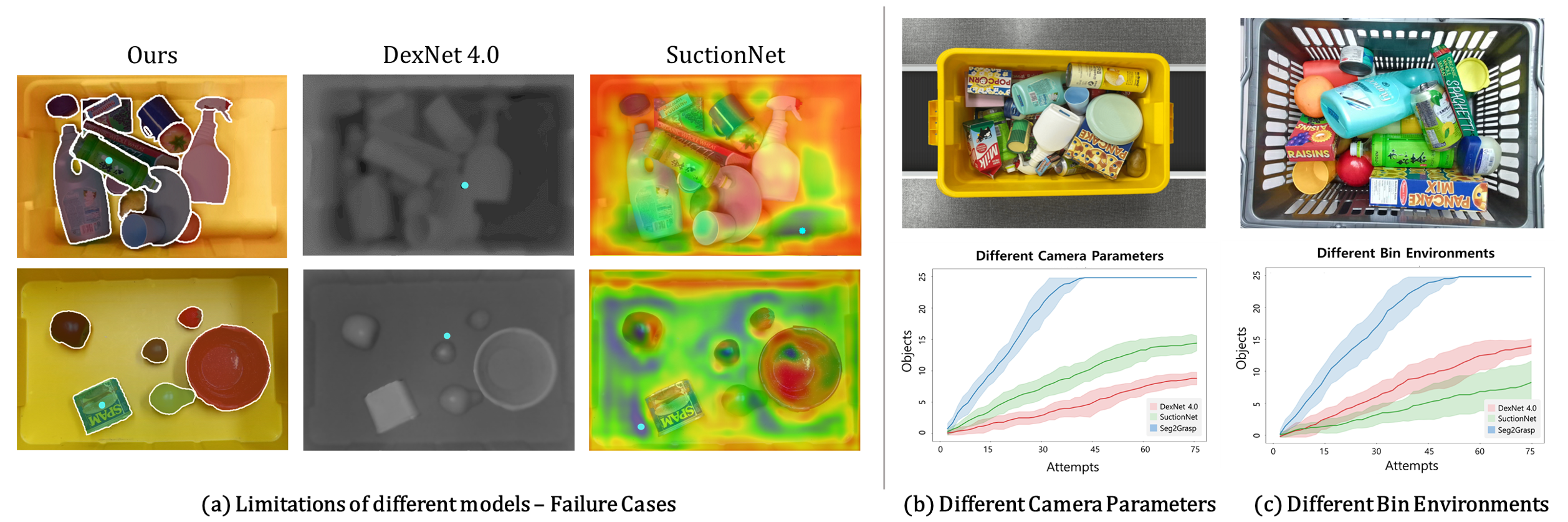}}
\vspace{-2mm}
\caption{\textbf{Experimental results.} (a) Failure cases highlighting the limitations of different models. (b) Setup and outcomes with different camera parameters. (c) Setup and outcomes with different bin environments.}
\label{result_graph}
\vspace{-1mm}
\end{figure*}

\textbf{Evaluation Metrics.}
Our bin picking task focuses on achieving the complete removal of items from a bin, evaluated through three principal metrics: The pick success rate, $pr = \frac{N_g}{N_t}$, quantifies the effectiveness of grasps, where $N_g$ represents the number of successful grasps and $N_t$ the total number of attempts. The object success rate, $or = \frac{N_g}{N_o}$, measures the accuracy of object handling, with $N_g$ indicating the number of objects successfully handled and $N_o$ the total count of objects. The segmentation success rate, $sr = \frac{N_s}{N_o}$, assesses the precision of segmentation, where $N_s$ signifies successfully segmented items, accurate within a tolerance of $\pm20\%$ of the ground truth area. Together, these critical metrics facilitate a comprehensive evaluation of our system's performance.

%% 실험 세팅 테이블

\subsubsection{Optimal Conditions Experiments}
We conducted experiments under conditions optimized for each method, ensuring that the environment was adjusted to provide the best possible performance. In this idealized setting, the objects and scenarios used in the experiments closely matched the data on which each model was trained. This approach allowed us to fairly assess the capabilities of each system in an environment that reflects its training conditions.

The evaluation was conducted across three levels of object distribution and complexity, progressively increasing the challenge to each model, as presented in Tab.~\ref{table_result}. In the simplest conditions, all methods performed comparably, with Seg2Grasp achieving a $pr$ of 89\%, slightly higher than DexNet 4.0 at 85\% and SuctionNet at 72\%. As the complexity increased, Seg2Grasp's performance advantage became more pronounced. Under the most challenging conditions, Seg2Grasp maintained a $pr$ of 79\%, significantly outperforming DexNet 4.0 at 28\% and SuctionNet at 29\%.

Similarly, Seg2Grasp excelled in object handling and segmentation accuracy. It achieved an $or$ of 86\% and a $sr$ of 83\% in the hardest scenarios, whereas DexNet 4.0 and SuctionNet showed considerable drops in performance, with $or$ and $sr$ both falling below 31\%.

% Dexnet 4.0's weakness, primarily its inability to consider object overlap. Consequently, it fails to grasp objects effectively when they are piled with heavier items. On the other hand, SuctionNet's downfall is its environment and object specificity, leading to repeating erroneous identification of the yellow background as a suction point, resulting in a bottleneck in the process. This problems are illustrated in Fig. \ref{result_graph}. (a). 

Both DexNet 4.0 and SuctionNet exhibit notable weaknesses under certain conditions. DexNet 4.0 struggles particularly with object overlap, making it difficult to grasp objects when heavier items are stacked on top. On the other hand, SuctionNet is highly sensitive to its environment and the specific objects it encounters, often misidentifying the yellow background as a suction point, which creates significant bottlenecks in the process. These issues are illustrated in Fig.~\ref{result_graph} (a).

These results underscore Seg2Grasp's superior robustness and adaptability, particularly under complex and variable conditions. The ability to maintain high accuracy across different levels of difficulty highlights its potential for real-world applications where precision and adaptability are critical.

\subsubsection{Varying Camera Parameters Experiments}
To evaluate the adaptability of the models under varying visual conditions, we conducted experiments focusing on the impact of different camera heights. The camera heights were altered to 50 cm (optimal), 100 cm, and 150 cm to assess how each model handles significant variations in the depth data, which are crucial for reliable object detection and grasping.

Over 10 experimental sets, each model was allowed up to 75 attempts per set to pick objects from the bin. As depicted in Fig.~\ref{result_graph}(b), SuctionNet and DexNet 4.0 showed considerable performance degradation at higher camera positions. DexNet 4.0, in particular, struggled due to increased noise in depth perception, successfully picking only 10 to 15 objects. SuctionNet fared slightly better but still demonstrated a significant drop in performance. In contrast, Seg2Grasp maintained high performance across all tested heights, showcasing remarkable resilience and consistency. These results highlight Seg2Grasp's robustness in adapting to changes in camera parameters, a critical factor in real-world robotic applications where environmental conditions can vary.

\begin{table} [t!]    
\centering
\caption{Results of Optimal Conditions Experiments Across Three Levels of Difficulty.}
\label{table_result}
\resizebox{\columnwidth}{!}{%
\begin{tabular}{lcccccc}
\toprule
\textbf{Level} & \makecell{\textbf{Setup} \\ (Layer, Item)} & \textbf{Method} & $pr$ & $or$ & $sr$ \\
\midrule
\multirow{3}{*}{\textbf{EASY}} & \multirow{3}{*}{Single, Trained} 
              & DexNet 4.0~\cite{dexnet40} & 0.85 & 0.94 & - \\
              & & SuctionNet~\cite{cao2021suctionnet} & 0.72 & 0.92 & 0.93 \\
              & & Ours & 0.89 & 0.96 & 0.91 \\
\midrule
\multirow{3}{*}{\textbf{MEDIUM}} & \multirow{3}{*}{Double, Mixed} 
              & DexNet 4.0~\cite{dexnet40} & 0.41 & 0.61 & - \\
              & & SuctionNet~\cite{cao2021suctionnet} & 0.51 & 0.53 & 0.43 \\
              & & Ours & 0.87 & 0.91 & 0.89 \\
\midrule
\multirow{3}{*}{\textbf{HARD}} & \multirow{3}{*}{Complex, Novel} 
              & DexNet 4.0~\cite{dexnet40} & 0.28 & 0.31 & - \\
              & & SuctionNet~\cite{cao2021suctionnet} & 0.29 & 0.23 & 0.26 \\
              & & Ours & 0.79 & 0.86 & 0.83 \\
\bottomrule
\end{tabular}%
}
\vspace{-3mm}
\end{table}

\subsubsection{Different Bin Environments Experiments}
We also examined the models' performance across different bin types to further test their adaptability. Starting with a large yellow bin as the baseline, we varied the bin types, including a shopping basket and a small white box, to simulate different real-world conditions. Each experimental set included 25 objects, none of which were encountered during training, with a maximum of 75 attempts allowed per set.

The outcomes, as shown in Fig.~\ref{result_graph}(c), reveal that both SuctionNet and DexNet 4.0 faced significant challenges when the bin type was changed. DexNet 4.0's reliance on depth data proved advantageous in scenarios with consistent lighting and minimal RGB variations, slightly outperforming SuctionNet in certain cases. However, neither model succeeded in fully clearing the bin under all conditions. On the other hand, Seg2Grasp consistently achieved high success rates regardless of the bin type, underscoring its superior adaptability and effectiveness in diverse environments. This consistency reinforces Seg2Grasp's capability to handle varying operational settings, making it a robust solution for industrial bin picking tasks.

\subsection{Open-Vocabulary Classification Experiment}
We conducted an experiment to compare the performance of Mask-CLIP against the standard CLIP model using our `mask-category' evaluation dataset. This involved computing CLIP feature vectors for evaluation mask images with both models and calculating their cosine similarity to 53 predefined categories.

The results, summarized in Tab.~\ref{maskclip_result}, indicate that Mask-CLIP outperforms the standard CLIP model in open-vocabulary classification tasks for mask-based inputs, achieving a Top-1 accuracy of 73.8\% and a Top-3 accuracy of 84.7\%, compared to CLIP's 66.1\% and 78.9\%, respectively.

\begin{table} [t!]
\renewcommand{\arraystretch}{1.2}
  \centering
  \caption{Results of the open-vocabulary classification.}\label{maskclip_result}
  \begin{tabular}{lccc}
  \toprule
  METHOD & Top-1 Acc. & Top-3 Acc. \\
  \midrule
  CLIP~\cite{radford2021learning} & 66.1\% & 78.9\% \\
  MASK-CLIP~\cite{liang2023open} & 73.8\% & 84.7\% \\
\bottomrule
% \vspace{-8mm}
\end{tabular}
\end{table}

\section{CONCLUSION}
This paper presented Seg2Grasp, a modular pipeline for enhancing suction grasping in dynamic, unstructured bin environments. The approach consists of three key modules: (1) Class-Agnostic Object Segmentation, isolating class-agnostic objects; (2) Grasp Pose Estimation using Surface Normals, identifying optimal suction points; and (3) Open-Vocabulary Object Classification, utilizing fine-tuned Mask-CLIP models for accurate identification. Real-robot experiments demonstrated that Seg2Grasp outperforms existing end-to-end bin picking models in both robustness and accuracy, highlighting the effectiveness of a modular strategy for automated bin picking.


%%%%%%%%%%%%%%%%%%%%%%%%%%%%%%%%%%%%%%%%%%%%%%%%%%%%%%%%%%%%%%%%%%%%%%%%%%%%%%%%

%%%%%%%%%%%%%%%%%%%%%%%%%%%%%%%%%%%%%%%%%%%%%%%%%%%%%%%%%%%%%%%%%%%%%%%%%%%%%%%%



% \section*{ACKNOWLEDGMENT}

% The preferred spelling of the word ÒacknowledgmentÓ in America is without an ÒeÓ after the ÒgÓ. Avoid the stilted expression, ÒOne of us (R. B. G.) thanks . . .Ó  Instead, try ÒR. B. G. thanksÓ. Put sponsor acknowledgments in the unnumbered footnote on the first page.

%%%%%%%%%%%%%%%%%%%%%%%%%%%%%%%%%%%%%%%%%%%%%%%%%%%%%%%%%%%%%%%%%%%%%%%%%%%%%%%%

% References are important to the reader; therefore, each citation must be complete and correct. If at all possible, references should be commonly available publications.

\bibliographystyle{IEEEtran}
\bibliography{ref}

\end{document}